\documentclass[letterpaper, 10 pt, conference]{ieeeconf}  

\IEEEoverridecommandlockouts                              

\usepackage[american]{babel}
\usepackage{graphicx} %
\usepackage{textcomp}
\usepackage{times} %
\usepackage{amsmath} %
\usepackage{amssymb}  %
\usepackage{siunitx} 
\usepackage{tikz}
\usepackage[ruled,vlined,noend, linesnumbered]{algorithm2e}
\usepackage{cite}
\usepackage{hyperref}
\usepackage[capitalise]{cleveref}
\AtBeginDocument{%
	\crefname{section}{Sec.}{Sec.}%
    \Crefname{Section}{Sec.}{Sec.}%
}
\usepackage{tikz}
\usepackage{pgfplots}
\pgfplotsset{compat=1.5.1,}
\usepackage{pgfplotstable}
\usepackage{standalone}
\usepackage[caption=false]{subfig}
\usepackage{booktabs}
\usepackage{multirow}
\usepackage[nohyperlinks, nolist]{acronym}
\usepackage{printlen}
\usepackage{tikzit}

\tikzstyle{red_circle}=[fill={rgb,255: red,227; green,27; blue,35}, draw=black, shape=circle, tikzit category=simple]
\tikzstyle{default_box}=[fill=white, draw={rgb,255: red,204; green,204; blue,204}, shape=rectangle, tikzit category=google, rounded corners=2mm, minimum width=2cm, minimum height=0.7cm, line width=0.4mm, text width=2cm, align=center, font={\footnotesize}]
\tikzstyle{higher_box}=[fill=white, draw={rgb,255: red,204; green,204; blue,204}, shape=rectangle, tikzit category=google, rounded corners=2mm, minimum width=2cm, minimum height=1.6cm, line width=0.4mm, text width=2cm, align=center, font={\footnotesize}]
\tikzstyle{tighter_box}=[fill=white, draw={rgb,255: red,204; green,204; blue,204}, shape=rectangle, tikzit category=google, rounded corners=2mm, minimum width=1cm, minimum height=0.7cm, line width=0.4mm, text width=1cm, align=center, font={\footnotesize}]
\tikzstyle{max_size}=[fill=none, draw=black, shape=rectangle, tikzit category=simple, minimum width=86.35902mm, minimum height=90mm, font={\footnotesize}]
\tikzstyle{wide_box}=[fill=white, draw={rgb,255: red,204; green,204; blue,204}, shape=rectangle, tikzit category=google, rounded corners=2mm, minimum width=2.5cm, minimum height=0.7cm, line width=0.4mm, text width=2.5cm, align=center, font={\footnotesize}]
\tikzstyle{wider_box}=[fill=white, draw={rgb,255: red,204; green,204; blue,204}, shape=rectangle, tikzit category=google, rounded corners=2mm, minimum width=3cm, minimum height=0.7cm, line width=0.4mm, text width=3cm, align=center, font={\footnotesize}]
\tikzstyle{extra_wide_box}=[fill=white, draw={rgb,255: red,204; green,204; blue,204}, shape=rectangle, tikzit category=google, rounded corners=2mm, minimum width=3.5cm, minimum height=0.7cm, line width=0.4mm, text width=3.5cm, align=center, font={\footnotesize}]
\tikzstyle{dashed_wide_box}=[fill=white, draw={rgb,255: red,204; green,204; blue,204}, shape=rectangle, tikzit category=google, dashed, rounded corners=2mm, minimum width=2.5cm, minimum height=0.7cm, line width=0.4mm, text width=2.5cm, align=center, font={\footnotesize}]
\tikzstyle{dashed_wider_box}=[fill=white, draw={rgb,255: red,204; green,204; blue,204}, shape=rectangle, tikzit category=google, dashed, rounded corners=2mm, minimum width=3cm, minimum height=0.7cm, line width=0.4mm, text width=3cm, align=center, font={\footnotesize}]
\tikzstyle{dashed_extra_wide_box}=[fill=white, draw={rgb,255: red,204; green,204; blue,204}, shape=rectangle, tikzit category=google, dashed, rounded corners=2mm, minimum width=3.5cm, minimum height=0.7cm, line width=0.4mm, text width=3.5cm, align=center, font={\footnotesize}]
\tikzstyle{large_content_box}=[fill=none, draw={rgb,255: red,204; green,204; blue,204}, shape=rectangle, tikzit category=google, rounded corners=2mm, minimum width=7.6cm, minimum height=3.5cm, line width=0.4mm, align=right, text width=7cm, text depth=3.5cm, font={\footnotesize}]
\tikzstyle{small_content_box}=[fill=none, draw={rgb,255: red,204; green,204; blue,204}, shape=rectangle, tikzit category=google, rounded corners=2mm, minimum width=7.6cm, minimum height=1.2cm, line width=0.4mm, align=left, text width=7cm, text depth=0.8cm, font={\footnotesize}]
\tikzstyle{image_box}=[fill=white, draw={rgb,255: red,204; green,204; blue,204}, shape=rectangle, tikzit category=google, line width=0.4mm, inner sep=0.2mm, outer sep=0pt, font={\footnotesize}]

\tikzstyle{left1_box}=[fill={rgb,255: red,234; green,119; blue,119}, draw=none, shape=rectangle, tikzit category=google, rounded corners=2mm, minimum width=2.0cm, minimum height=1.9cm, line width=0.4mm, align=right, text width=2.0cm, text depth=1.3cm, font={\footnotesize}]
\tikzstyle{left2_box}=[fill={rgb,255: red,234; green,119; blue,119}, draw=none, shape=rectangle, tikzit category=google, minimum width=2.5cm, minimum height=1.9cm, line width=0.0mm, align=right, text width=2.5cm, text depth=1.3cm, font={\footnotesize}]
\tikzstyle{right2_box}=[fill={rgb,255: red,109; green,158; blue,235}, draw=none, shape=rectangle, tikzit category=google, rounded corners=2mm, minimum width=3.5cm, minimum height=1.9cm, line width=0.4mm, align=left, text width=3.5cm, text depth=1.3cm, font={\footnotesize}]
\tikzstyle{right1_box}=[fill={rgb,255: red,109; green,158; blue,235}, draw=none, shape=rectangle, tikzit category=google, minimum width=0.8cm, minimum height=1.9cm, line width=0.0mm, align=left, text width=0.8cm, text depth=1.3cm, font={\footnotesize}]

\tikzstyle{my_arrow}=[tikzit category=google, draw={rgb,255: red,204; green,204; blue,204}, line width=0.4mm, ->, >={Triangle[scale=1]}]
\tikzstyle{my_blue_arrow}=[tikzit category=google, draw={rgb,255: red,109; green,158; blue,235}, line width=0.4mm, ->, >={Triangle[scale=1]}]
\tikzstyle{my_red_arrow}=[tikzit category=google, draw={rgb,255: red,234; green,119; blue,119}, line width=0.4mm, ->, >={Triangle[scale=1]}]
\tikzstyle{my_line}=[-, tikzit category=google, draw={rgb,255: red,204; green,204; blue,204}, line width=0.4mm]
\tikzstyle{my_blue_line}=[-, tikzit category=google, draw={rgb,255: red,109; green,158; blue,235}, line width=0.4mm]
\tikzstyle{my_red_line}=[-, tikzit category=google, draw={rgb,255: red,234; green,119; blue,119}, line width=0.4mm]
\tikzstyle{shield_edge}=[-, line width=0.6mm, draw={rgb,255: red,204; green,204; blue,204}, fill={rgb,255: red,220; green,220; blue,20}, tikzit category=google, dashed]

\usetikzlibrary{arrows.meta}
\usetikzlibrary{fit}
\definecolor{TUMblue}{rgb}{0.00, 0.40, 0.74}
\definecolor{TUMdarkblue}{rgb}{0.00, 0.32, 0.576}
\definecolor{TUMlightblue}{rgb}{0.392, 0.627, 0.784}
\definecolor{TUMlighterblue}{rgb}{0.596, 0.776, 0.917}
\definecolor{TUMgray}{rgb}{0.6, 0.6, 0.6}
\definecolor{TUMlightgray}{rgb}{0.855, 0.843, 0.796}
\definecolor{TUMorange}{rgb}{0.89, 0.447, 0.133}
\definecolor{TUMgreen}{rgb}{0.635, 0.678, 0.00}
\definecolor{mygreen}{RGB}{0, 146, 0}

\SetCommentSty{mycommfont}

\newcommand{\Hrgym}{\texttt{Human-robot gym} }
\newcommand{\hrgym}{\texttt{human-robot gym} }
\newcommand{\hrgyM}{\texttt{human-robot gym}}
\newcommand{\robosuite}{robosuite }
\newcommand{\robosuitE}{robosuite}

\usepackage{amsmath,amsfonts,bm}

\def\eqref#1{equation~\ref{#1}}

\def\1{\bm{1}}

\def\rvn{{\mathbf{n}}}

\def\ervn{{\textnormal{n}}}

\def\va{{\bm{a}}}

\def\vd{{\bm{d}}}

\def\vo{{\bm{o}}}
\def\vp{{\bm{p}}}
\def\vq{{\bm{q}}}

\def\vs{{\bm{s}}}

\def\vx{{\bm{x}}}

\DeclareMathAlphabet{\mathsfit}{\encodingdefault}{\sfdefault}{m}{sl}
\SetMathAlphabet{\mathsfit}{bold}{\encodingdefault}{\sfdefault}{bx}{n}

\def\sA{{\mathcal{A}}}
\def\sB{{\mathcal{B}}}

\def\sS{{\mathcal{S}}}

\newcommand{\R}{\mathbb{R}}

\newcommand{\smid}{\, | \,}

\newcommand{\expertk}{\ensuremath{\pi_\mathrm{e}(\va_k \smid \vs_k)}}

\newcommand{\noisyexpertk}{\ensuremath{\tilde{\pi}_\mathrm{e}(\va_k \smid \vs_k, k)}}
\DeclareMathOperator{\dist}{dist}

\title{\LARGE \bf
Human-Robot Gym: Benchmarking Reinforcement Learning in Human-Robot Collaboration
}

\author{Jakob Thumm, Felix Trost, and Matthias Althoff
\thanks{The authors are with the Department of Computer Engineering, Technical University of Munich, Germany.
        {\tt\small jakob.thumm@tum.de}, {\tt\small felix.trost@tum.de}, {\tt\small althoff@tum.de}\newline
    \textcopyright 2024 IEEE.  Personal use of this material is permitted.  Permission from IEEE must be obtained for all other uses, in any current or future media, including reprinting/republishing this material for advertising or promotional purposes, creating new collective works, for resale or redistribution to servers or lists, or reuse of any copyrighted component of this work in other works.}%
}

\begin{document}
\begin{acronym}
\acro{rl}[RL]{reinforcement learning}
\acro{mdp}[MDP]{Markov decision process}
\acro{ppo}[PPO]{proximal policy optimization}
\acro{sac}[SAC]{soft actor-critic}
\acro{hrc}[HRC]{human-robot collaboration}
\acro{dof}[DoF]{degree of freedom}
\acro{iss}[ISS]{invariably safe state}
\acro{ssm}[SSM]{speed and separation monitoring}
\acro{pfl}[PFL]{power and force limiting}
\acro{rsi}[RSI]{reference state initialization}
\acro{sir}[SIR]{state-based imitation reward}
\acro{air}[AIR]{action-based imitation reward}
\acro{pid}[PID]{proportional–integral–derivative}
\acro{pdf}[PDF]{probability density function}
\end{acronym}

\maketitle
\thispagestyle{empty}
\pagestyle{empty}

\begin{abstract}
Deep reinforcement learning (RL) has shown promising results in robot motion planning with first attempts in human-robot collaboration (HRC).
However, a fair comparison of RL approaches in HRC under the constraint of guaranteed safety is yet to be made.
We, therefore, present {\normalfont\hrgyM}, a benchmark suite for safe RL in HRC. 
We provide challenging, realistic HRC tasks in a modular simulation framework.
Most importantly, {\normalfont\hrgym}is the first benchmark suite that includes a safety shield to provably guarantee human safety.
This bridges a critical gap between theoretic RL research and its real-world deployment.
Our evaluation of six tasks led to three key results: (a) the diverse nature of the tasks offered by {\normalfont\hrgym} creates a challenging benchmark for state-of-the-art RL methods, (b) by leveraging expert knowledge in form of an action imitation reward, the RL agent can outperform the expert, and (c) our agents negligibly overfit to training data.
\end{abstract}
\section{Introduction}\label{sec:introduction}
Recent advancements in deep \ac{rl} are promising for solving intricate decision-making processes~\cite{brohan_2022_RT1Robotics} and complex manipulation tasks~\cite{liu_2021_DeepReinforcementa}. 
These capabilities are essential for \ac{hrc}, given that robotic systems must act in environments featuring highly nonlinear human dynamics.
Despite the promising outlook, the few works on \ac{rl} in \ac{hrc} confine themselves to narrow task domains~\cite{semeraro_2023_HumanRobot}.
Two primary challenges impeding the widespread integration of \ac{rl} in \ac{hrc} are safety concerns and the diversity of tasks.
The assurance of safety for \ac{rl} agents operating within human-centric environments is a hurdle as agents generate potentially unpredictable actions, posing substantial risks to human collaborators.
Current \ac{hrc} benchmarks~\cite{erickson_2020_AssistiveGym, ye_2022_RCareWorldHumancentric} circumvent these safety concerns by focusing on interacting with primarily stationary humans.

In this paper, we propose \hrgyM\footnote{\hrgym is available at \url{https://github.com/TUMcps/human-robot-gym}}, a suite of \ac{hrc} benchmarks that comes with a broad range of tasks, including object inspection, handovers, and collaborative manipulation, while ensuring safe robot behavior by integrating \mbox{SaRA shield~\cite{thumm_2022_ProvablySafe}}, a tool for provably safe \ac{rl} in \ac{hrc}.
With its set of challenging \ac{hrc} tasks, \hrgym enables training \ac{rl} agents to collaborate with humans in a safe manner, which is not possible with other benchmarks.
\Hrgym comes with pre-defined benchmarks that are easily extendable and adjustable.
We then track all relevant performance and safety metrics to allow an extensive evaluation of the solutions.
\begin{figure}[t]
	\centering
	\vspace{1.7mm}
	\hspace*{\fill}%
	\begin{minipage}[t]{0.245\linewidth}
		\vspace{0pt}
		\includegraphics[width=21.15mm, height=19.5mm, trim={25cm 17cm 22cm 2cm}, clip]{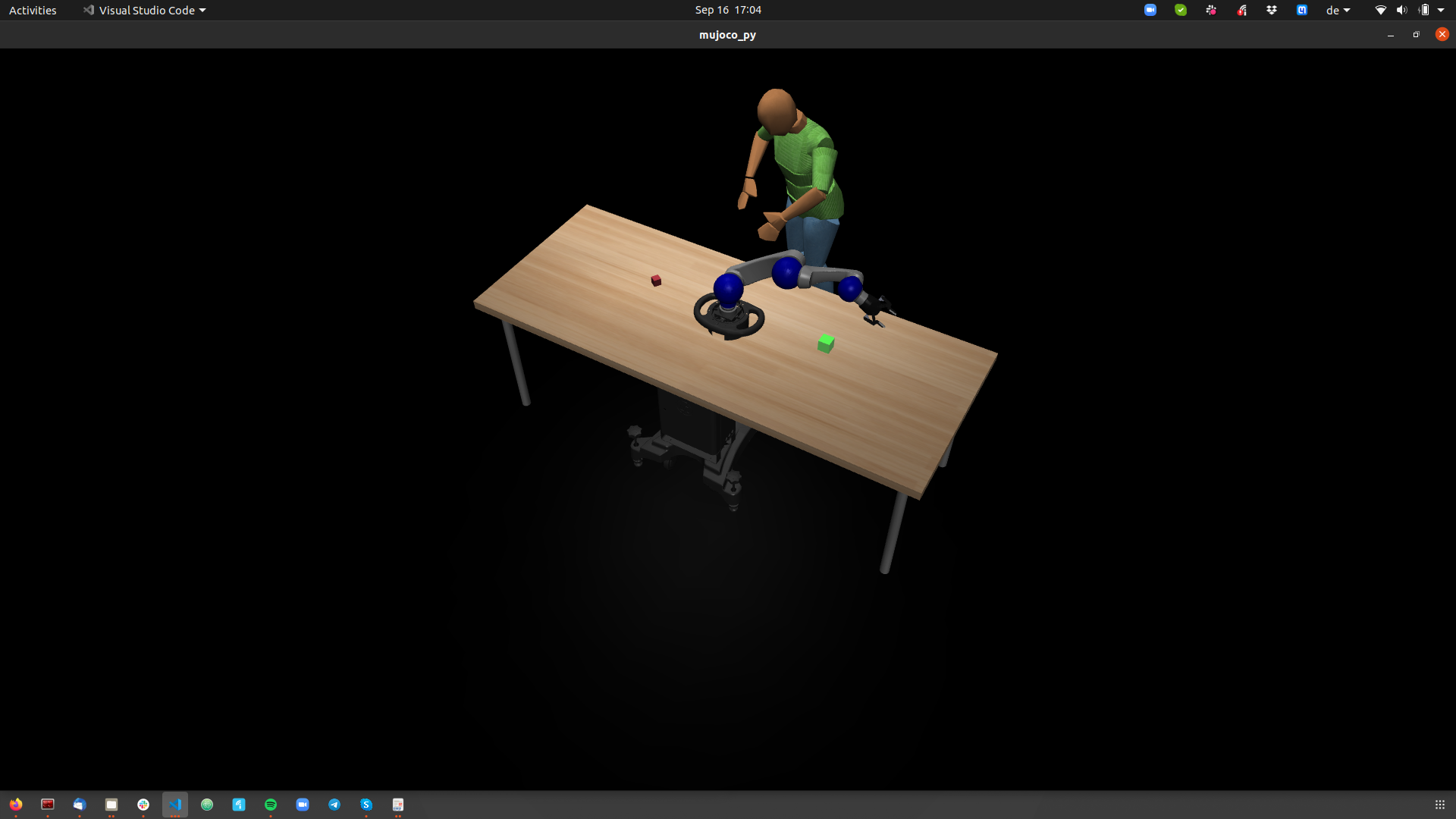}
		\centering
		\footnotesize{Reach}
	\end{minipage}
	\hspace{-5pt}
	\begin{minipage}[t]{0.245\linewidth}
		\vspace{0pt}
		\includegraphics[width=21.15mm, height=19.5mm, trim={28cm 14cm 19cm 5cm}, clip]{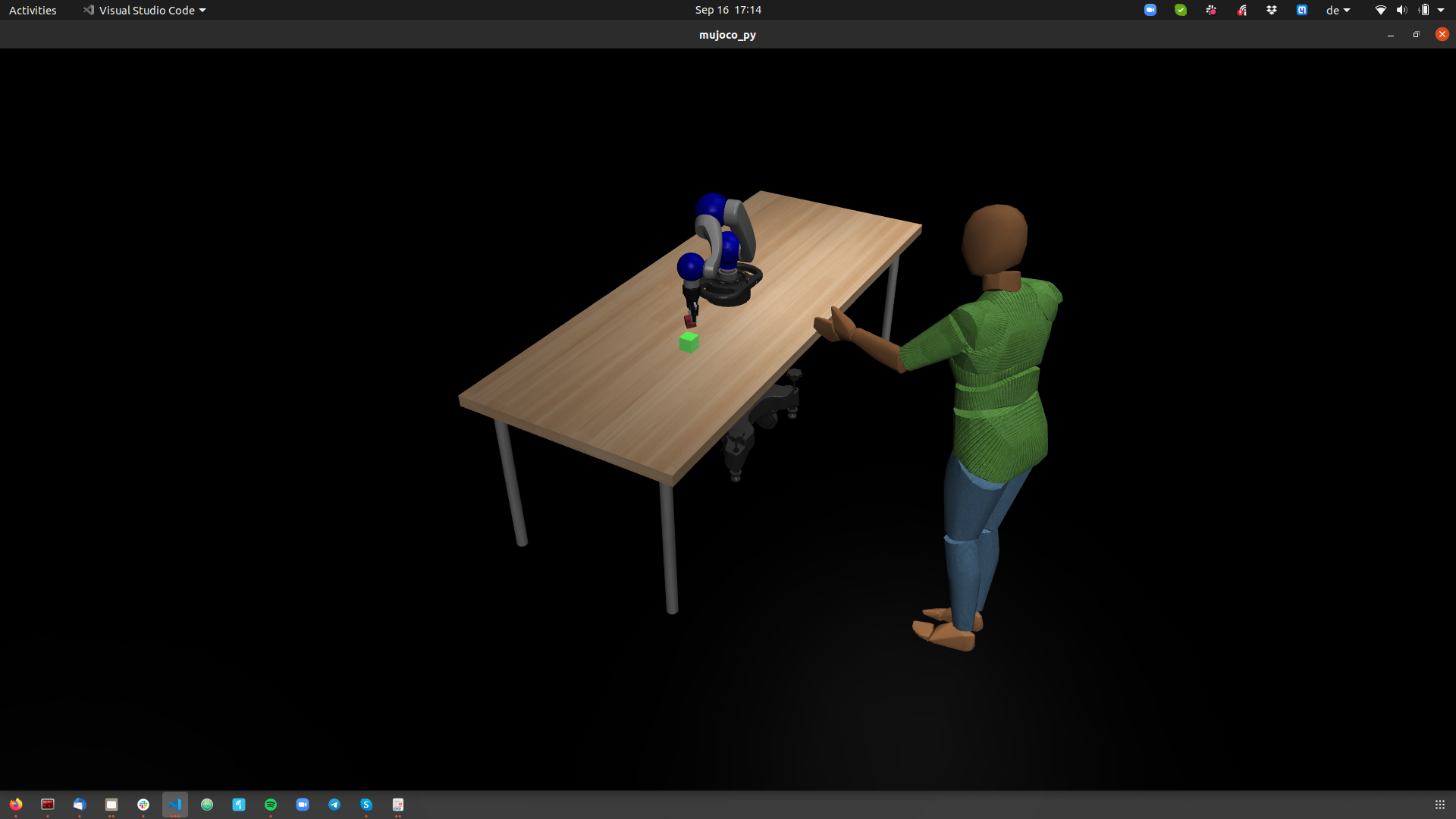}
		\centering
		\footnotesize{Pick and place}
	\end{minipage}
	\hspace{-5pt}
	\begin{minipage}[t]{0.245\linewidth}
		\vspace{0pt}
		\includegraphics[width=21.15mm, height=19.5mm, trim={12cm 10.5cm 29cm 2.5cm}, clip]{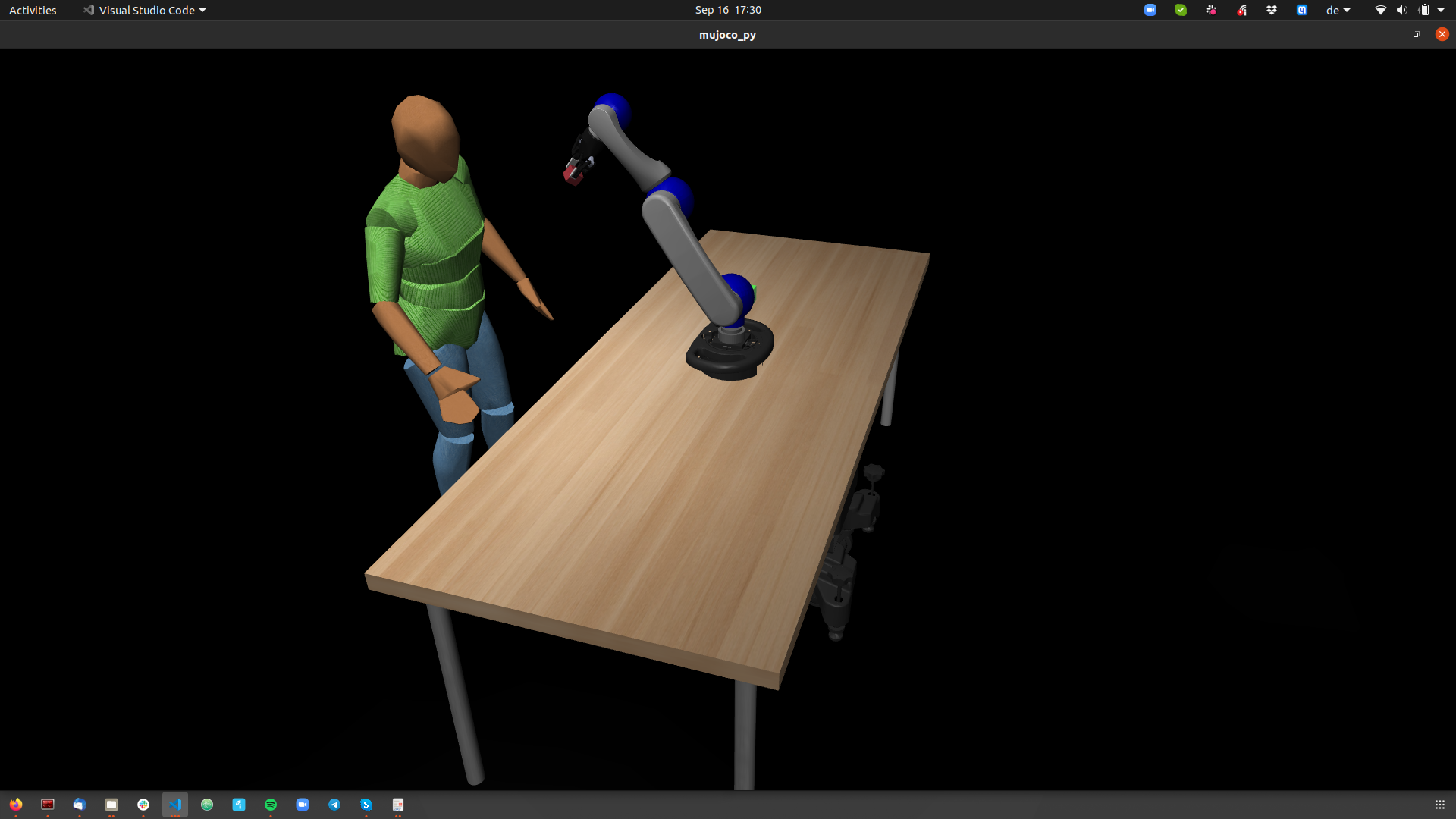}
		\centering
		\footnotesize{Object inspection}
	\end{minipage}
	\hspace{-5pt}
	\begin{minipage}[t]{0.245\linewidth}
		\vspace{0pt}
		\includegraphics[width=21.15mm, height=19.5mm, trim={21cm 16cm 26cm 3cm}, clip]{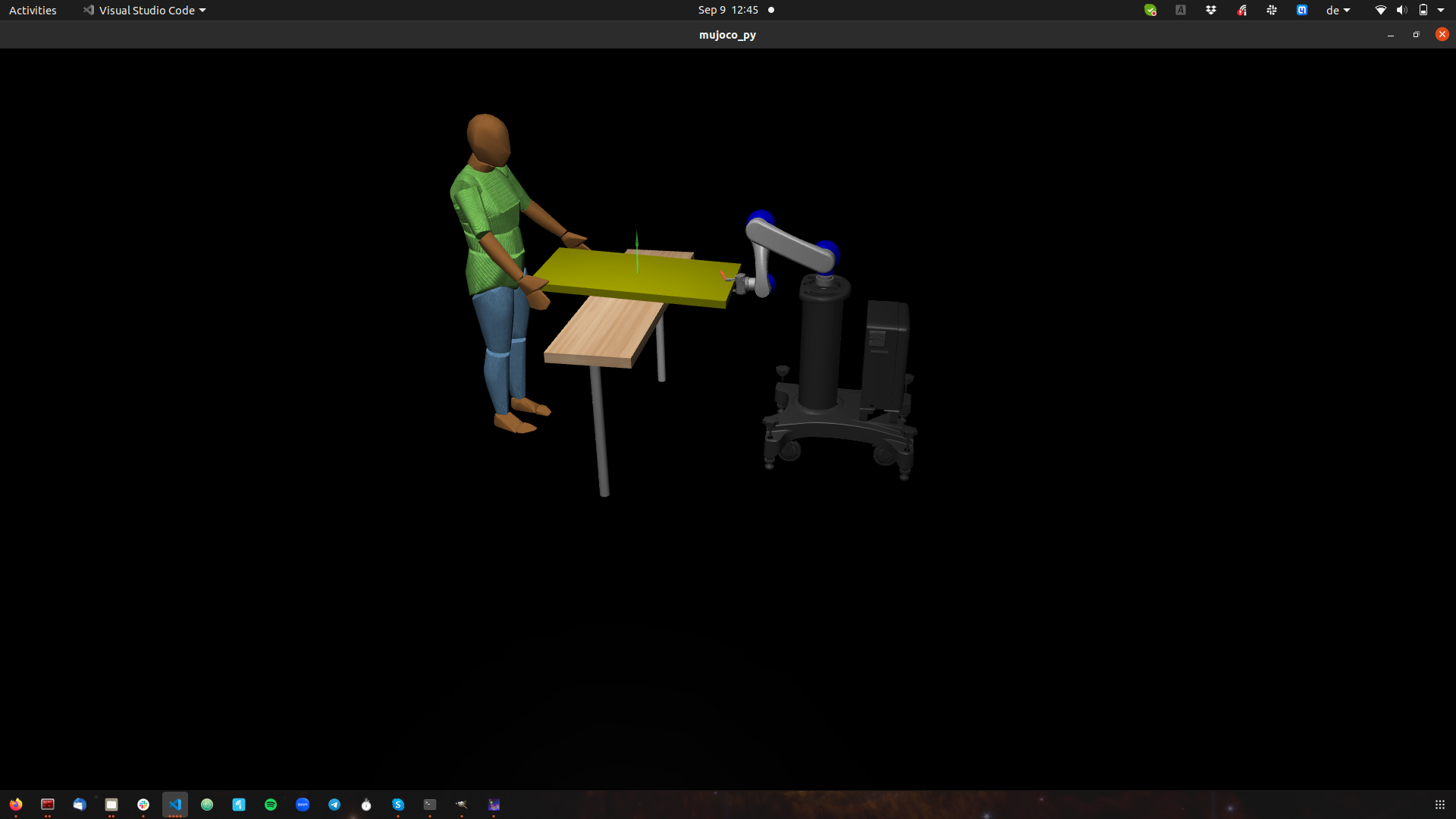}
		\centering
		\footnotesize{Lifting}
	\end{minipage}
	\hspace*{\fill}\\[-0.9em]
	\hspace*{\fill}%
	\begin{minipage}[t]{0.245\linewidth}
		\vspace{0pt}
		\includegraphics[width=21.15mm, height=19.5mm, trim={28cm 15cm 19cm 4cm}, clip]{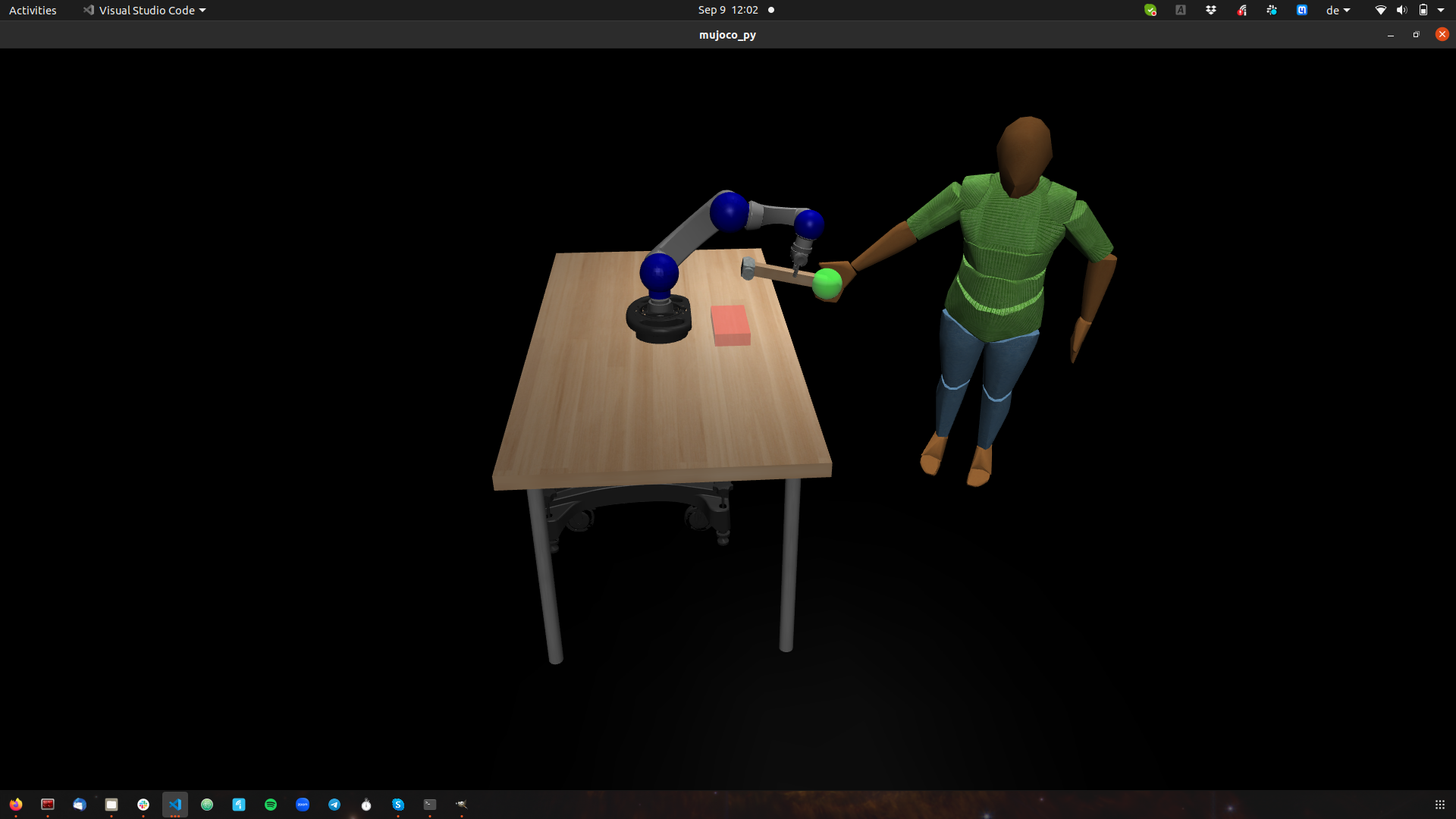}
		\centering
		\footnotesize{Robot-human handover}
	\end{minipage}
	\hspace{-5pt}
	\begin{minipage}[t]{0.245\linewidth}
		\vspace{0pt}
		\includegraphics[width=21.15mm, height=19.5mm, trim={28cm 15cm 19cm 4cm}, clip]{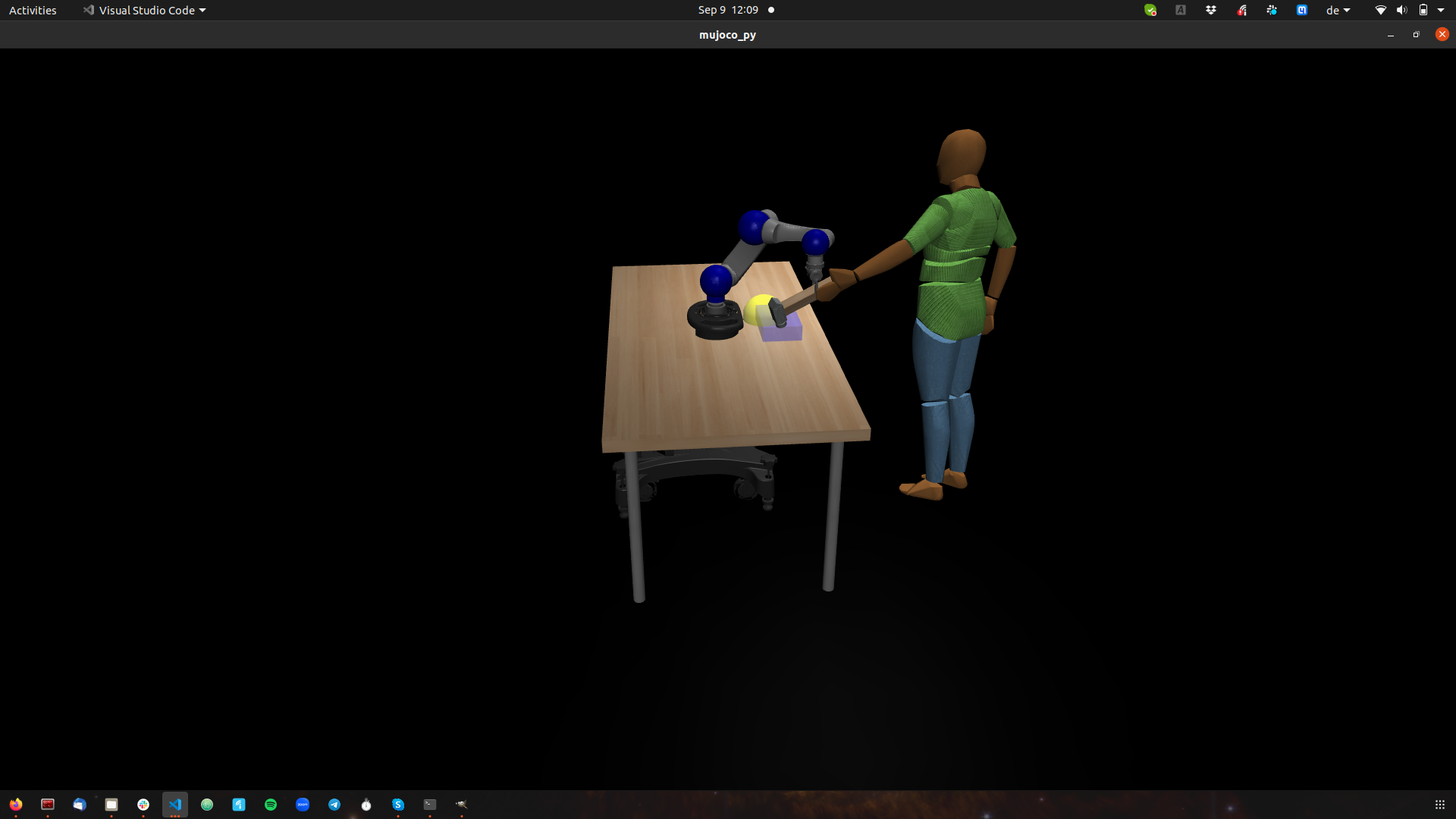}
		\centering
		\footnotesize{Human-robot handover}
	\end{minipage}
	\hspace{-5pt}
	\begin{minipage}[t]{0.245\linewidth}
		\vspace{0pt}
		\includegraphics[width=21.15mm, height=19.5mm, trim={19cm 11.5cm 26cm 6.0cm}, clip]{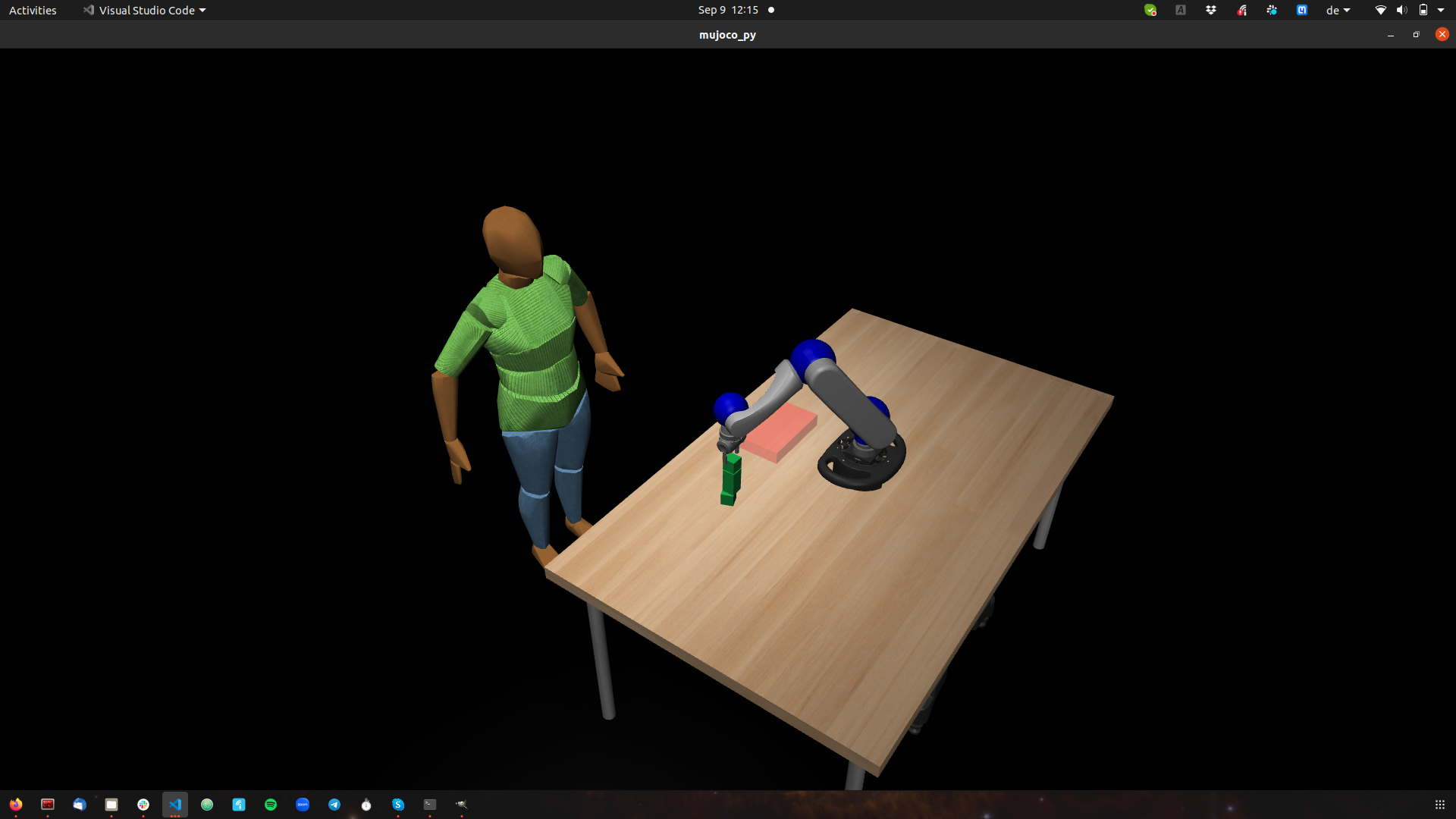}
		\centering
		\footnotesize{Collaborative stacking}
	\end{minipage}
	\hspace{-5pt}
	\begin{minipage}[t]{0.245\linewidth}
		\vspace{0pt}
		\includegraphics[width=21.15mm, height=19.5mm, trim={30cm 16cm 17cm 3cm}, clip]{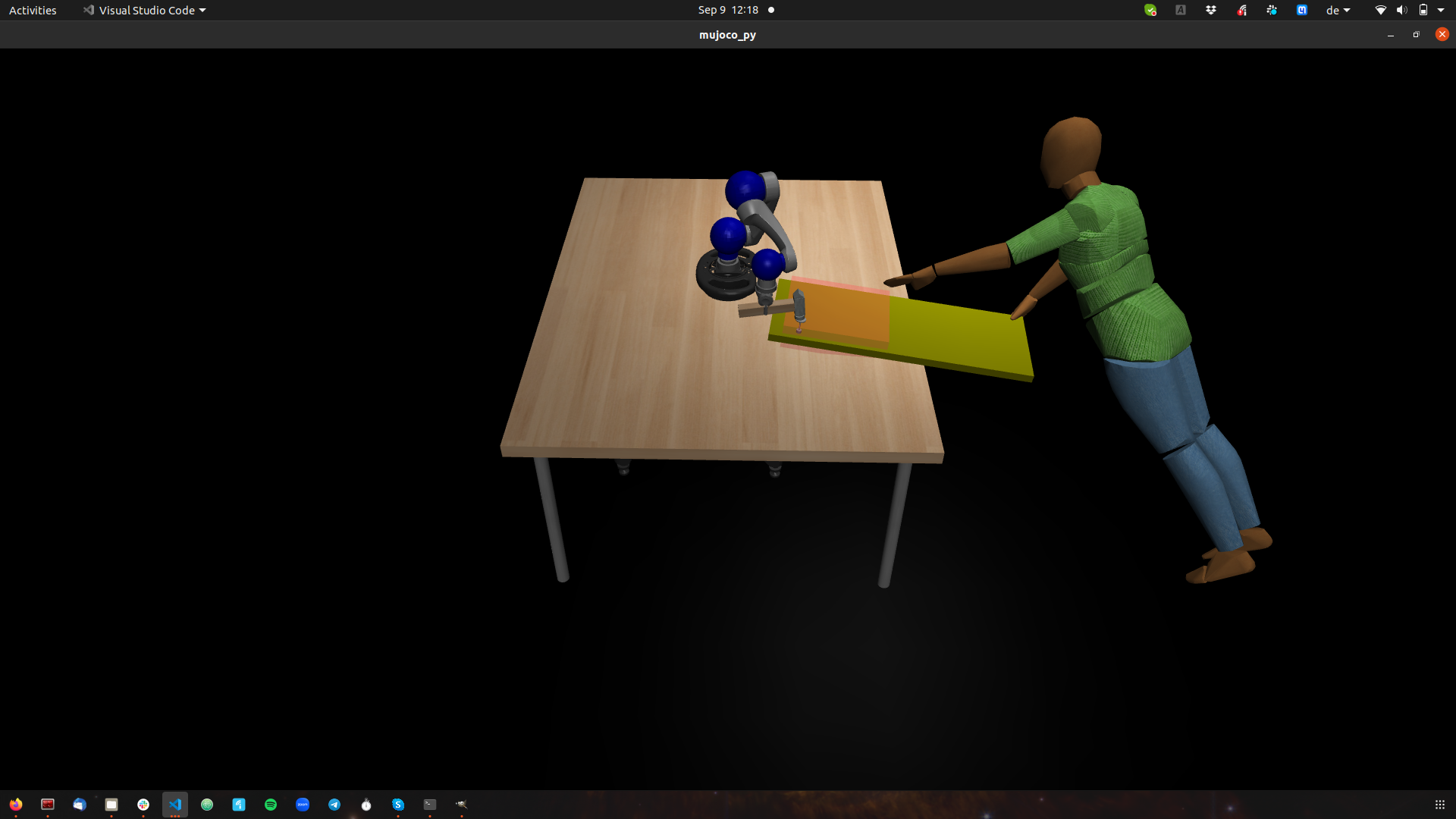}
		\centering
		\footnotesize{Collaborative hammering}
	\end{minipage}
	\hspace*{\fill}%
	\caption{\Hrgym presents eight challenging \ac{hrc} tasks.}
	\label{fig:environments}
\end{figure}
Our benchmark suite features the following key elements that lower the entry barrier into the field of \ac{rl} in \ac{hrc}:
\begin{itemize}
	\item Pre-defined tasks, see~\cref{fig:environments}, with varying difficulty, each with a set of real-world human movements.
    \item Available robots: Panda, Sawyer, IIWA, Jaco, Kinova3, UR5e, and Schunk.
	\item Provable safety for \ac{hrc} using SaRA shield in addition to static and self-collision prevention.
	\item High fidelity simulation based on MuJoCo~\cite{todorov_2012_MuJoCoPhysics}.
    \item Support of joint space and workspace actions.
	\item Highly configurable and expandable benchmarks. 
    \item Environment definition based on the OpenAI gym standard to support state-of-the-art \ac{rl} frameworks, such as stable-baselines 3~\cite{raffin_2021_Stablebaselines3Reliable}.
	\item Pre-defined expert policies for gathering imitation data and performance comparison.
    \item Easily reproducible baseline results, see \cref{sec:experiments}.
\end{itemize}
This article is structured as follows: 
\cref{sec:related_work} introduces previous work in \ac{rl} for \ac{hrc}, compares \hrgym to other related benchmarks in the field, and gives a short overview of imitation learning approaches.
\cref{sec:benchmark} presents our benchmark suite in detail.
We then present additional tools supporting users to solve \hrgym tasks in \cref{sec:tools}.
\cref{sec:experiments} evaluates our benchmarks experimentally and discusses the results.
Finally, we conclude this work in \cref{sec:conclusion}. 
\section{Related Work}\label{sec:related_work}
Semeraro et al.~\cite{semeraro_2023_HumanRobot} summarize recent efforts in machine learning for \ac{hrc}.
They identify four typical \ac{hrc} applications: collaborative assembly \cite{vogt_2017_SystemLearninga, cunha_2020_CollaborativeRobots}, object handover \cite{maeda_2017_ProbabilisticMovement, shukla_2018_LearningSemantics, lagomarsino_2023_MaximisingCoefficiency}, object handling \cite{roveda_2020_ModelBasedReinforcement, deng_2017_HierarchicalRobot}, and collaborative manufacturing~\cite{nikolaidis_2015_EfficientModel}.

Recent developments in \ac{rl} evoke the need for comparable benchmarks in various applications.
One of the most used benchmark suites for robotic manipulation is \robosuitE \cite{zhu_2020_RobosuiteModular}, which offers a set of diverse robot models, realistic sensor and actuator models, simple task generation, and a high-fidelity simulation using MuJoCo~\cite{todorov_2012_MuJoCoPhysics}.
Further notable manipulation benchmarks are included in Orbit~\cite{mittal_2023_OrbitUnified}, which focuses on photorealism; Behavior-1K~\cite{li_2022_BEHAVIOR1KBenchmark}, which provides 1000 everyday robotic tasks in the simulation environment OmniGibson; and meta-world~\cite{yu_2020_MetaworldBenchmark} for meta \ac{rl} research.

None of the benchmarks mentioned above include humans in the simulation.
There are, however, some benchmarks that provide limited human capabilities with a specific research focus.
First, the robot interaction in virtual reality \cite{higgins_2021_MakingVirtual} and SIGVerse~\cite{inamura_2021_SIGVerseCloudbased} benchmarks include real humans in real-time teleoperation through virtual reality setups.
Unfortunately, this approach is unsuitable for training an \ac{rl} agent from scratch due to long training times.
Closest to our work are AssistiveGym~\cite{erickson_2020_AssistiveGym} and RCareWorld~\cite{ye_2022_RCareWorldHumancentric}. 
These benchmark suites provide simulation environments for ambulant caregiving tasks. 
RCareWorld provides a large set of assistive tasks using a realistic human model and a choice of robot manipulators.
However, AssistiveGym and RCareWorld focus on tasks where the human is primarily static or only features small, limited movements. 
Comparably, our work focuses on collaborative tasks, where the human and the robot play an active role, and the human movement is thus complex.
Furthermore, one primary focus of \mbox{\hrgym} is human safety, which other benchmarks only cover superficially.
Also closely related to our work is HandoverSim~\cite{chao_2022_HandoverSimSimulation}, which investigates the handover of diverse objects from humans to robots.
Here, prerecorded motion-capturing clips steer the human hand. 
However, these movements only capture the hand picking up objects and presenting them to the robot. 
From that point onward, the hand remains motionless~\cite{chao_2022_HandoverSimSimulation}. 
Compared to our work, HandoverSim (a) does not supply motion data while the handover is ongoing, (b) has a much narrower selection of tasks, and (c) excludes safety concerns.

We utilize learning from experts~\cite{ramirez_2022_ModelfreeReinforcement} to provide the first results on our benchmarks.
Currently, we mainly rely on two techniques: \acl{rsi}, which lets the agent start at a random point of an expert trajectory \cite{uchendu_2023_JumpStartReinforcement}, and \acl{sir}, which additionally rewards the agent for being close to the expert trajectory \cite{peng_2018_DeepMimicExampleguided}.
We explicitly decided against behavior cloning techniques~\cite{zheng_2022_ImitationLearning} as they merely copy the expert behavior and often fail to generalize to the task objective~\cite{ramirez_2022_ModelfreeReinforcement}.

\section{Benchmark suite}\label{sec:benchmark}
\begin{figure}
	\centering
	\vspace{1.7mm}
	\includegraphics[scale=1]{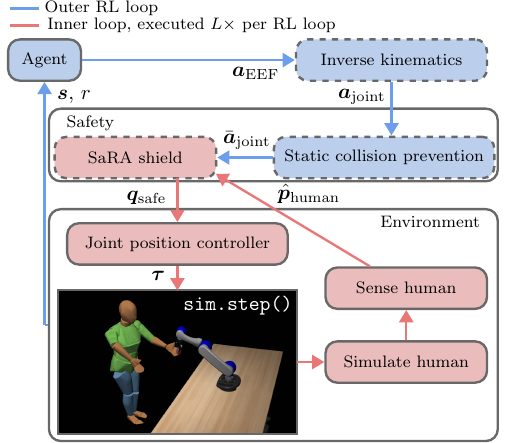}
	\caption{
		A typical workflow of an \ac{rl} cycle in \hrgyM. Optional elements are depicted with dashed borders, and the inner loop of the environment step is executed $L$ times, e.g., $L=25$. In this example, the agent returns an action in Cartesian space corresponding to a desired end effector position, which is converted to a desired joint position using inverse kinematics. Our collision prevention alters the action if the desired joint position results in a self-collision or a collision with the static environment. The shield calculates the next safe joint positions, which the joint position controller converts into joint torques that are then executed in simulation.
	}
	\label{fig:main_structure}
\end{figure}

We base \hrgym on \robosuite \cite{zhu_2020_RobosuiteModular}, which already provides adjustable robot controllers and a high-fidelity simulation environment with MuJoCo.
Primarily, our environment introduces the functionality to interact with a human entity, define tasks with complex collaboration objectives, and evaluate human safety.
In the following subsections, we describe our benchmarks, the typical workflow of \hrgyM, and its elements in more detail.

\subsection{Benchmark definition}
We define a benchmark in \hrgym by its robot ($\mathcal{R}$), reward\footnote{We can convert our rewards into costs used in~\cite{mayer_2024_CoBRAComposable} by $c = -r$.} ($\mathcal{C}$), and task ($\Theta$) following the definition in~\cite[Eq. 1]{mayer_2024_CoBRAComposable}.
All our benchmarks are described by modular configuration files via the Hydra framework~\cite{yadan_2019_HydraFramework}, which makes \hrgym easily configurable and extendable.
Each benchmark has a main configuration file, consisting of pointers to configuration files for the task and reward definition, robot specifications, environment wrapper settings, expert policy descriptions, training parameters, and \ac{rl} algorithm hyperparameters.

\paragraph{Robot} We currently support seven different robot models: Panda, Sawyer, IIWA, Jaco, Kinova3, UR5e, and Schunk.

\paragraph{Reward} The reward in our environments can be sparse, e.g., indicating whether an object is at the target position, and dense, e.g., proportional to the Euclidean distance of the end effector to the goal. 
Furthermore, environments can have a delayed sparse reward signal, which should mimic a realistic \ac{hrc} environment, where the agent receives the task fulfillment reward shortly after the action that completes the task.
An example of a delayed reward is when a handover was successful, but the human needs a short time to approve the execution.
The reward delay serves as an additional challenge for the \ac{rl} agents.

\paragraph{Task definition} 
Each task in \hrgym is defined by a safety mode, objects, obstacles, human motions, and a set of goals, adding to the task definition of~\cite{mayer_2024_CoBRAComposable}.
\Hrgym features tasks that reflect the \ac{hrc} categories introduced in~\cite{semeraro_2023_HumanRobot}.
Additionally, we selected two typical coexistence tasks: reach as well as pick and place.
Furthermore, we provide a pipeline to generate new human movements from motion capture data, which allows users to define their own tasks and extend \hrgyM.
\Cref{tab:environments} displays the default settings of each task that we use in our experiments, and a subjective estimate of the authors on the relative difficulty of each task regarding manipulation, length of the time horizon, and human dynamics.
The details of the safety modes are discussed in~\cref{sec:safety}.

\begin{table*}
	\vspace{1.7mm}
	\centering
	\caption{Benchmark characteristics}
	\begin{scriptsize}
		\begin{tabular}{lcccccccc} 
			\toprule
			\textbf{Task} & \textbf{\ac{hrc} category}$^{1}$ & \textbf{Safety mode}$^2$ & \textbf{Manipulation} & \textbf{Time-horizon} & \textbf{Dynamics} & \textbf{Reward} & \textbf{Reward delay} & \textbf{No. of motions}\\
			\midrule 
			Reach & coexistence & \acs{ssm} & easy & easy & easy & dense & no & 12\\
			Pick and place & coexistence & \acs{ssm} & medium & medium & easy & sparse & no & 12\\
			Object inspection & object handling & \acs{ssm} & medium & medium & medium & sparse & yes & 8\\
			Collaborative lifting & object handling & \acs{ssm} & medium & medium & medium & dense & no & 9\\
			Robot-human handover & object handover & \acs{pfl} & medium & medium & hard & sparse & yes & 15\\
			Human-robot handover & object handover & \acs{pfl} & hard & medium & hard & sparse & yes & 11\\
			Collaborative hammering & object manufacturing & \acs{ssm} & hard & medium & hard & sparse & yes & 11\\
			Collaborative stacking & object assembly & \acs{ssm} & hard & hard & hard & sparse & yes & 8\\
			\bottomrule
			~\\[-7pt]
			\multicolumn{9}{l}{$^1$from~\cite{semeraro_2023_HumanRobot}, $^2$SSM: \acl{ssm}, PFL: \acl{pfl}}
		\end{tabular}
	\end{scriptsize}
	\label{tab:environments}
\end{table*}

\subsection{Typical workflow}
\cref{fig:main_structure} displays a typical workflow of an \ac{rl} cycle in \hrgyM.
The actions of the \ac{rl} agent can be joint space $\va_{\text{joint}}$ or workspace actions $\va_{\text{EEF}}$.
If $\va_{\text{EEF}}$ is selected, the inverse kinematics wrapper determines $\vp_{\text{joint, desired}}$ from $\vp_{\text{EEF, desired}}$ and returns the joint action.
The workspace actions can include the end effector orientation in $\mathrm{SO}(3)$.
However, in our experiments, we only use the desired positional difference of the end effector in Cartesian space $\va_{\text{EEF}} = \vp_{\text{EEF, desired}} - \vp_{\text{EEF}}$ as actions, where the gripper is pointing downwards.
This simplification to a four-dimensional action space (three positional actions and a gripper action) is common in literature~\cite{liu_2021_DeepReinforcementa, andrychowicz_2017_HindsightExperience, li_2020_PracticalMultiObject}.
Training in joint space showed similar performance in first experiments but required significantly more \ac{rl} steps until convergence due to the larger action space.

The \ac{rl} action might violate safety constraints.
Users can, therefore, implement safety functionalities as part of the outer \ac{rl} loop or the inner environment loop.
We present how our additional tools use both variants to prevent collisions with static obstacles and guarantee human safety in~\cref{sec:tools}.
The step function of our environment executes its inner loop $L$ times.
Every iteration of the inner loop runs the optional inner safety function, the robot controller, one fixed step of the MuJoCo simulation, and the human measurement.
After executing the action, the environment returns an observation and a reward to the agent. 

\subsection{Human simulation}
Our simulation moves the human using motion capture files obtained from a Vicon tracking system.
All movements are recorded specifically for the defined tasks and include task-relevant objects in the scene, ensuring realistic behavior.
A limitation of using recordings are instances where the recording must be paused until the robot initiates a specific event, e.g., in a handover task.
Previous works show an unnatural human behavior in these cases.
To address this limitation, we incorporate idle movements representing the human waiting for an event to trigger.
For each recording, keyframes can designate the start and end of an idle phase.
Once reached, the movement remains idle until an event predicate $\sigma_E$ is true, at which point it progresses to the successive movement.
The predicate $\sigma_E$ is true when the robot achieves a task-specific sub-goal and thereafter, e.g., handing over an object. 
Instead of simply looping the idle phase, which would lead to jumps in the movement, we alter the replay time of the recording by a set of $D$ superimposing sine-functions: 
\begin{align}
	t_A = 
	\begin{cases}
		&t, \text{ if } t \leq t_I \lor \sigma_E  \\
		&t_I + \sum\limits_{i=1}^D \upsilon_i \sin \left(\left(t-t_I\right) \omega_i\right), \text{ otherwise} \, ,
	\end{cases}
\end{align}
where $\upsilon_i$ and $\omega_i$ define the amplitude and frequency of the $i$-th sine-function during idling respectively and both are randomized at the start of each episode.
The replay time can also reverse in the idling phase.
The recordings to replay are randomly selected at the start of each episode, and their starting position and orientation are slightly randomized to avoid overfitting.

\subsection{Observation}
\begin{table}[t]
	\centering
	\caption{Observation elements}
	\label{tab:observations}
	\setlength{\tabcolsep}{4.5pt}
	\begin{scriptsize}
		\begin{tabular}{c|lc} 
			\toprule
			& \textbf{Element} & \textbf{Observations}\\
			\midrule
			\multirow{4}{*}{\rotatebox{90}{Robot ($\mathcal{R}$)}} & Joint angle & $\vq$\\
			& Joint velocity & $\dot{\vq}$\\
			& EEF aperture & $\varphi$\\
			& EEF pose & $\vp_{\text{W}}(\mathbf{T}_{\text{EEF}})^\dagger$, $\vo_{\text{W}}(\mathbf{T}_{\text{EEF}})$\\
			\midrule
			\multirow{10}{*}{\rotatebox{90}{Task ($\Theta$)}}& Objects & $\vp_{\text{W}}(\mathbf{T}_{\text{obj}})$, $\vp_{\text{E}}(\mathbf{T}_{\text{obj}})$, $\vd(\mathbf{T}_{\text{obj}})$, $\vo_{\text{W}}(\mathbf{T}_{\text{obj}})$\\
			& Obstacles & $\vp_{\text{W}}(\mathbf{T}_{\text{obs}})$, $\vp_{\text{E}}(\mathbf{T}_{\text{obs}})$, $\vd(\mathbf{T}_{\text{obs}})$, $\vo_{\text{W}}(\mathbf{T}_{\text{obs}})$\\
			& Goal poses & $\vp_{\text{W}}(\mathbf{T}_{\text{goal}})$, $\vp_{\text{E}}(\mathbf{T}_{\text{goal}})$, $\vd(\mathbf{T}_{\text{goal}})$, $\vo_{\text{W}}(\mathbf{T}_{\text{goal}})$\\
			& Goal joint angles & $\vq_{\text{goal}} - \vq$\\
			& Object gripped & $\sigma_{\text{grip}}$\\
			& Object at target & $\sigma_{\text{target}}$\\
			& Static collision & $\sigma_{\text{col, stat}}$\\
			& Body positions & $\vp_{\text{W}}(\mathbf{T}_{\text{body}})$, $\vp_{\text{E}}(\mathbf{T}_{\text{body}})$, $\vd(\mathbf{T}_{\text{body}})$\\
			& Safe human contact & $\sigma_{\text{contact}}$\\
			& Critical human contact & $\sigma_{\text{crit}}$\\
			\bottomrule
		\end{tabular} 
	\end{scriptsize}
    \newline
    \vspace{2mm}
	\tiny{$^\dagger \vp$: position in world (W) or end effector (E) frame, $\vd$: Euclidean distance, $\vo$: orientation, $\sigma$: predicate}
\end{table}

\Hrgym features typical task-related and robotic observations, as shown in~\cref{tab:observations}.
Objects, obstacles, goals, and human bodies have a measurable pose \mbox{$\mathbf{T} \in SE(3)$}. 
These objects are observable through the following projections (adapted from~\cite[Tab. II]{mayer_2024_CoBRAComposable}): position in world (W) and end effector (E) frame \mbox{$\vp_{\text{W}} : SE(3) \rightarrow \R^3$}, \mbox{$\vp_{\text{E}} : SE(3) \rightarrow \R^3$}, Euclidean distance to the end effector \mbox{$\vd : SE(3) \rightarrow \R^+$}, and the orientation in world frame given through quaternions \mbox{$\vo_{\text{W}} : SE(3) \rightarrow SO(3)$}.
The task-specific elements in~\cref{tab:observations} include those necessary to fulfill the task, i.e., \mbox{$\mathbf{T}_{\text{obj},a}, a = 1, \dots, A$}, \mbox{$\mathbf{T}_{\text{obs},b}, b = 1, \dots, B$}, \mbox{$\mathbf{T}_{\text{goal},c}, c = 1, \dots, C$}, and \mbox{$\mathbf{T}_{\text{body},d}, d = 1, \dots, D$}, with $A$ objects, $B$ obstacles, $C$ goal poses, and $D$ human bodies.
The robot information contains its joint positions and velocities as well as the end effector position, orientation, and aperture.
In our experiments, we found that reducing the number of elements in the observation, e.g., only providing measurements of the human hand positions instead of the entire human model, is beneficial for training performance.
To emulate real-world sensors, users can optionally add noise sampled from a compact set and delays to all measurements, further reducing the gap between simulation and reality.
In addition to the physical measurements, the user can define cameras that observe the scene and learn from vision inputs.

\section{Supporting tools}\label{sec:tools}
This section describes additional tools included in \hrgym to provide safety and \ac{rl} training functionality.
\subsection{Safety tools}\label{sec:safety}
We can prevent static and self-collisions in the outer \ac{rl} loop by performing collision checks of the desired robot trajectory using pinocchio~\cite{carpentier_2019_PinocchioLibrary}.
If the trajectory resulting from the \ac{rl} action is unsafe, we sample actions uniformly from the action space until we find a safe action.

Guaranteeing human safety in the outer \ac{rl} loop is challenging, as the time horizon of \ac{rl} actions is relatively long, e.g., \SI{200}{\milli \second}.
Hence, checking safety only once before execution would lead to a very restrictive safety behavior~\cite{thumm_2023_ReducingSafety}.
Therefore, we ensure human safety in the inner environment loop.
We provide the tool SaRA shield introduced for robotic manipulators in~\cite{althoff_2019_EffortlessCreation, thumm_2022_ProvablySafe} and generalized to arbitrary robotic systems in \cite{thumm_2023_ReducingSafety}.
First, SaRA shield translates each \ac{rl} action into an intended trajectory. 
In the subsequent period of an \ac{rl} action, the shield is executed $L$ times.
In each timestep, the shield computes a failsafe trajectory, which guides the robot to an \acl{iss}.
As defined in~\cite{thumm_2022_ProvablySafe}, an \acl{iss} in manipulation is a condition where the robot completely stops in compliance with the ISO~10218-1~2021 regulations~\cite{iso_2021_RoboticsSafety}.
Next, the shield constructs a shielded trajectory combining one timestep from the planned intended trajectory with the failsafe trajectory.  
SaRA shield validates these shielded trajectories through set-based reachability analysis of the human and robot.
For this, the shield receives the position and velocity of human body parts as measurements from the simulation.
We assure safety indefinitely, provided that the initial state of the system is an \acl{iss}, by only executing the step from the intended trajectory when the shielded trajectory is confirmed safe~\cite{thumm_2022_ProvablySafe}.
In the event of a failed safety verification, the robot follows the most recently validated failsafe trajectory, guaranteeing continued safe operation.
Finally, SaRA shield returns the desired robot joint states for the next timestep to follow the verified trajectory.
We then use a \acl{pid} controller to calculate the desired robot joint torques.

The default mode of SaRA shield is \acl{ssm}, which stops the robot before an imminent collision.
This is too restrictive for close interaction tasks, such as handovers, as the robot must come into contact with the human.
Therefore, we include a \acl{pfl} mode in the tool SaRA shield that decelerates the robot to a safe Cartesian velocity of \SI{5}{\milli\meter \per \second} before any human contact, as proposed in~\cite[Def.~3]{beckert_2017_OnlineVerification}.
Thereby, our \acl{pfl} mode ensures painless contact in accordance with ISO~10218-1~2021~\cite{iso_2021_RoboticsSafety}.
As in the \acl{ssm} mode, SaRA shield only slows down the robot if our reachability-based verification detects a potential collision.
Otherwise, the robot is allowed to operate at full speed.
We further plan to include a conformant impedance controller, as proposed in~\cite{liu_2021_OnlineVerification}, in SaRA shield in the future.

\subsection{Tools for training}
To provide a perspective on the performance of \ac{rl} agents in our environments, we provide both expert and \ac{rl} policies with our tasks.
In this work, we consider an \ac{rl} agent that learns on a \acl{mdp} described by the tuple $\left(\sS, \sA, T, r, \sS_0, \gamma\right)$ in both continuous or discrete action spaces $\sA$ and continuous state spaces $\sS$ with a set of initial states $\sS_0$.
Here, $T(\vs_{k+1} \smid \vs_k, \va_k)$ is the transition function, which denotes the \acl{pdf} of transitioning from state $\vs_k$ to $\vs_{k+1}$ when action $\va_k$ is taken. 
The agent receives a reward determined by the function \mbox{$r : \sS \times \sA \times \sS\rightarrow \mathbb{R}$} from the environment. 
Lastly, we consider a discount factor \mbox{$\gamma \in [0, 1]$} to adjust the relevance of future rewards.
\ac{rl} aims to learn an optimal policy $\pi^\star(\va_k \smid \vs_k)$ that maximizes the expected return $R = \sum_{k=0}^{K} \gamma^k r(\vs_k, \va_k, \vs_{k+1})$ when starting from an initial state $\vs_0 \in \sS_0$ and following $\pi^\star(\va_k \smid \vs_k)$ until termination at $k=K$~\cite{sutton_2018_ReinforcementLearning}.

\subsection{Pre-defined experts}
We define a deterministic expert policy $\expertk$ for each task to gather imitation data and compare performance.
The experts are hand-crafted and follow a proportional control law with heuristics based on human expertise strategy, as described in full detail in the \hrgym documentation.

To achieve diversity in our expert data, we add a noise term to the expert action, resulting in the noisy expert
\begin{align}\label{eq:expert_with_noise}
\noisyexpertk =  \expertk * f_{k, \rvn} \, ,
\end{align} 
where $*$ denotes the convolution of probability distributions, and $f_{k, \rvn}$ is the \acl{pdf} of the noise signal $\rvn$ at time $k$.
To restrain the random process from diverting too far from the expert, we choose a mean-reverting process.
In particular, we model $\rvn$ to be a vector of independent random variables $\ervn_{i}$ and discretize the univariate Ornstein--Uhlenbeck process~\cite{uhlenbeck_1930_TheoryBrowniana} to retrieve an autoregressive model of \mbox{order one}.
We can sample an expert trajectory \mbox{$\chi = \left(\tilde{\vs}_0, \dots, \tilde{\vs}_K\right)$} by a Monte Carlo simulation, where we start in \mbox{$\tilde{\vs}_0 \in \sS_0$}, and subsequently follow \mbox{$\tilde{\vs}_{k+1} \sim T(\tilde{\vs}_{k+1} \smid \tilde{\vs}_k, \tilde{\va}_k)$} with \mbox{$\tilde{\va}_k \sim \tilde{\pi}_\mathrm{e}(\tilde{\va}_k \smid \tilde{\vs}_k, k)$} for \mbox{$k = 0, \dots, K-1$}.
For each task in \hrgyM, we provide the expert policies $\pi_{\text{e}}$ and $\tilde{\pi}_{\text{e}}$ together with a set of $M$ expert trajectories $\sB = \left\{\chi_1, \dots, \chi_M\right\}$ sampled from $\tilde{\pi}_{\text{e}}$.

\subsection{Reinforcement learning agents}\label{sec:rl_agents}
\Ac{sac}~\cite{haarnoja_2018_SoftActorCritic} serves as a baseline for our experiments due to its sample efficiency and good performance on previous experiments~\cite{thumm_2022_ProvablySafe}.
We include three variants of imitation learning to investigate the benefit of expert knowledge for the \ac{rl} agent.
First, we use \acl{rsi}~\cite{peng_2018_DeepMimicExampleguided} to redefine the set of initial states to the set of states contained in the expert trajectories \mbox{$\sS_0 = \left\{ \tilde{\vs} \mid \tilde{\vs} \in \chi, \chi \in \sB\right\}$}.
Starting the episode from a state reached by the expert informs the agent about reachable states and their reward in long-horizon tasks.

Secondly, we evaluate a \acl{sir}, where the agent receives an additional reward signal proportional to its closeness to an expert trajectory \mbox{$\chi \in \sB$} in state space $r_{\text{SIR}}(\vs_k, \va_k, \vs_{k+1}, \tilde{\vs}_k) = (1-\varsigma) r(\vs_k, \va_k, \vs_{k+1}) + \varsigma \dist(\vs_k - \tilde{\vs}_k)$, where \mbox{$0 \le \varsigma \ll 1$}.
For the distance function, we choose a scaled Gaussian function $\dist(\vx) = 2^{-\kappa \|\vx\|_2}$ with scaling factor $\frac{1}{\kappa}$ as suggested in~\cite{peng_2018_DeepMimicExampleguided}.
We further apply \acl{rsi} when using the \acl{sir}, as proposed in~\cite{peng_2018_DeepMimicExampleguided}.

Finally, we adapt the \acl{sir} method to an \acl{air}, where the agent receives an additional reward signal proportional to the closeness of its action to the expert action $r_{\text{AIR}}(\vs_k, \va_k, \vs_{k+1}, \tilde{\va}_k) = (1-\varsigma) r(\vs_k, \va_k, \vs_{k+1}) + \varsigma \dist(\va_k - \tilde{\va}_k)$, with \mbox{$\tilde{\va}_k \sim \tilde{\pi}_\mathrm{e}(\tilde{\va}_k \smid \vs_k, k)$}.
When using \aclp{air}, we sample the expert policy alongside the \ac{rl} policy in every step but only execute the \ac{rl} action.
\section{Experiments}\label{sec:experiments}
This section presents the evaluated \ac{rl} agents, shows the performance of the agents in \hrgyM, and discusses the results.
Our experiments aim to answer three main research questions:
\begin{itemize}
    \item Can \ac{rl} be used to complete complex \ac{hrc} tasks?
    \item How beneficial is prior expert knowledge in solving these tasks?
    \item Does the \ac{rl} agent overfit to a limited amount of human recordings in training?
\end{itemize}

\begin{figure*}[ht]
	\vspace{1.7mm}
	\begin{minipage}{177.7mm}\centering
		\subfloat{
			\includegraphics[scale=1]{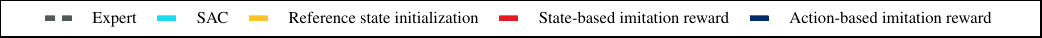}
		}
	\end{minipage}\\[3mm]
	\hspace*{8.7mm}
	\fboxsep0pt
	\colorbox{white}{\begin{minipage}{25.0mm}
			\centering
			\footnotesize{Reach}
	\end{minipage}}
	\hspace*{0.3mm}
	\colorbox{white}{\begin{minipage}{25.0mm}
			\centering
			\footnotesize{Pick and Place}
	\end{minipage}}
	\hspace*{0.3mm}
	\colorbox{white}{\begin{minipage}{25.0mm}
			\centering
			\footnotesize{Collaborative Lifting}
	\end{minipage}}
	\hspace*{0.3mm}
	\colorbox{white}{\begin{minipage}{25.0mm}
			\centering
			\footnotesize{Robot-Human Handover}
	\end{minipage}}
	\hspace*{0.3mm}
	\colorbox{white}{\begin{minipage}{25.0mm}
			\centering
			\footnotesize{Human-Robot Handover}
	\end{minipage}}
	\hspace*{0.3mm}
	\colorbox{white}{\begin{minipage}{25.0mm}
			\centering
			\footnotesize{Collaborative Stacking}
	\end{minipage}}
	\\[-3mm]
	\subfloat{
		\includegraphics[scale=1]{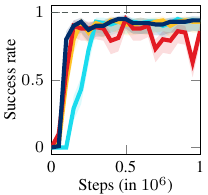}
	}\hspace*{-3.5mm}
	\subfloat{
		\includegraphics[scale=1]{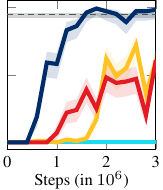}
	}\hspace*{-3.5mm}
	\subfloat{
		\includegraphics[scale=1]{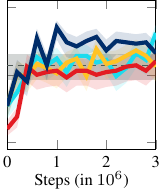}
	}\hspace*{-3.5mm}
	\subfloat{
		\includegraphics[scale=1]{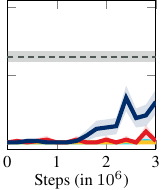}
	}\hspace*{-3.5mm}
	\subfloat{
		\includegraphics[scale=1]{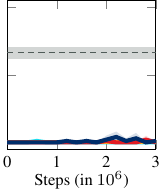}
	}\hspace*{-3.5mm}
	\subfloat{
		\includegraphics[scale=1]{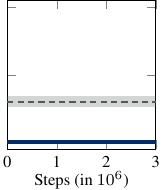}
	}\\[-0.1cm]
	\subfloat{
		\includegraphics[scale=1]{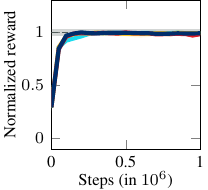}
	}\hspace*{-3.5mm}
	\subfloat{
		\includegraphics[scale=1]{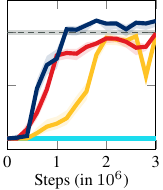}
	}\hspace*{-3.5mm}
	\subfloat{
		\includegraphics[scale=1]{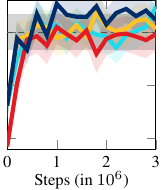}
	}\hspace*{-3.5mm}
	\subfloat{
		\includegraphics[scale=1]{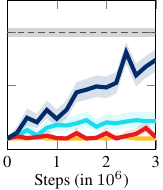}
	}\hspace*{-3.5mm}
	\subfloat{
		\includegraphics[scale=1]{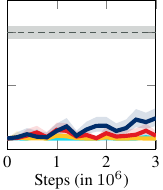}
	}\hspace*{-3.5mm}
	\subfloat{
		\includegraphics[scale=1]{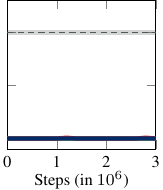}
	}\hspace*{-3.5mm}
	\caption{Evaluation performance during training of \ac{rl} agents on \hrgyM. The plots show the mean evaluation performance during training and the 95\% confidence interval in the mean metric obtained with bootstrapping when training on five random seeds.}
    \label{fig:main_results}
\end{figure*}
\begin{figure}[ht]
	\centering
	\includegraphics[scale=1]{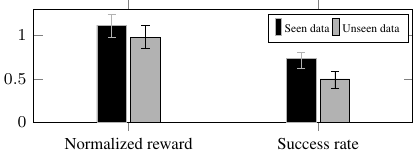}
	\caption{Ablation study for overfitting to motion data in the training process.}
	\label{fig:ablation_overfitting}
\end{figure}
We present our results on six \hrgym tasks: reach, pick and place, collaborative lifting, robot-human handover, human-robot handover, and collaborative stacking.
The evaluation shows results for the Schunk robot with the rewards listed in~\cref{tab:environments}.
Across all experiments, we execute $L=25$ safety shield steps per \ac{rl} step (empirically, training with $L=50$ shows similar performance).
Our training had an average runtime of \SI{17.61}{\second} per $10^3$ \ac{rl} steps\footnote{Run on ten cores of an AMD EPYC\textsuperscript{\texttrademark} 7763 @ 2.45GHz.}.
To evaluate the benefit of expert knowledge in these complex tasks, we compare the four agents discussed in~\cref{sec:rl_agents} with the expert.
\cref{fig:main_results} shows our main results, where we evaluate the performance every \num{2e5} training steps and trained all agents on five random seeds.
We report the success rate, which indicates the rate at which the task was successful, and the reward normalized to the range between the minimal possible reward and the average expert reward.
All plots show the mean evaluation performance during training and the \SI{95}{\%} confidence interval (shaded area) in the mean metric established with bootstrapping on $10^4$ samples.

Our results show that the \hrgym has a diverse set of tasks, from which some are already solvable, e.g., reach as well as pick and place, some show room for improvement, e.g., collaborative lifting and robot-human handover, and some are not solvable with the investigated approaches, e.g., human-robot handover and collaborative stacking.
Comparing these results to the complexity estimate in~\cref{tab:environments}, we infer that the two main factors for the difficulty of a task are the complexity of the manipulation and the human dynamics.
Handling these two areas will be among the main challenges for \ac{rl} research in \ac{hrc}.

The results in~\cref{fig:main_results} further show that expert knowledge is beneficial in benchmarks with sparse rewards, with the \ac{air} method showing higher or equal performance compared to the state-based one.
In the pick and place task, the \acl{air} approach outperformed the expert policy and reached a nearly \SI{100}{\%} success rate.
Unfortunately, constructing the \acl{air} requires an expert policy that can be queried online during training, which is not given in many manipulation tasks.
Interestingly, the agent trained with a \acl{sir} shows no significant improvement over the \ac{sac} agent trained only with \acl{rsi} in our evaluations.
Our results indicate that starting the environment in meaningful high-reward states significantly improves performance in sparse reward settings.
Future work could investigate if there are even more effective forms of \acl{rsi} that require little to no expert knowledge.
Finally, expert knowledge does not improve performance in our experiments with dense reward settings, such as reaching or collaborative lifting.
We assume this behavior stems from the fact that the additional action-based and state-based imitation rewards resemble the dense environment reward, yielding little additional information. 

To address concerns related to overfitting to the limited amount of human motion profiles, we conduct an ablation study on the collaborative lifting task, which relies exceedingly on the human motion.
This study aims to identify whether training an \ac{rl} agent using a limited set of recordings instead of simulated behavior is satisfactory. 
Our dataset consists of nine unique human motion captures, seven of which we use as training data, reserving the remaining two for testing.
We then perform a five-fold cross-evaluation, where we select different training and testing movements on each split and train \ac{rl} agents on five random seeds per split.
We report the average performance over the splits and seeds and the \SI{95}{\%} confidence interval in the mean metric of the trained \ac{sac} agent on the respective training movements (seen data) and test movements (unseen data) in~\cref{fig:ablation_overfitting}.
The reward performance of the trained agent on the unseen data is within the confidence interval of the performance on the training data.
Both mean reward and success rate are only slightly lower on the unseen data, and the agent performs reasonably well.
Therefore, we conclude that overfitting to the human movements is not a significant problem of \hrgyM.
\section{Conclusion}\label{sec:conclusion}
\Hrgym offers a realistic benchmark suite for comparing performance of \ac{rl} agents and safety functions in \ac{hrc}.
Its unique provision of a pre-implemented safety shield offers the opportunity to develop efficient \ac{hrc} without designing a safety function.
Our evaluation insights reveal the importance of expert knowledge in benchmarks with sparse rewards, showing that an \acl{air} is a promising approach if an expert is available online.
In terms of practical application, it is noteworthy that an agent trained in \hrgym was successfully deployed in actual \ac{hrc} environments, as presented in our prior work~\cite{thumm_2023_ReducingSafety}.
These tests underline the critical role \hrgym will play as an academic tool and as a practical approach for tangible robotic issues.

\section*{Acknowledgment}
The authors gratefully acknowledge financial support by
the Horizon 2020 EU Framework Project CONCERT under grant 101016007.
\addtolength{\textheight}{0cm}   



\bibliographystyle{IEEEtran}
\bibliography{library_cleaned}

\end{document}